%% file: main.tex
\documentclass[sigconf,natbib=true]{acmart}
\usepackage{algorithm, algorithmic}
\usepackage{amsmath}
\usepackage{commath}
\usepackage{mathtools}
\usepackage{nicefrac}
\usepackage{subfig}
\usepackage{balance}
\pdfoutput=1

\DeclarePairedDelimiter\floor{\lfloor}{\rfloor}

\AtBeginDocument{%
  \providecommand\BibTeX{{%
    \normalfont B\kern-0.5em{\scshape i\kern-0.25em b}\kern-0.8em\TeX}}}

\copyrightyear{2021} 
\acmYear{2021} 
\setcopyright{acmcopyright}\acmConference[CIKM '21]{Proceedings of the 30th ACM International Conference on Information and Knowledge Management}{November 1--5, 2021}{Virtual Event, QLD, Australia}
\acmBooktitle{Proceedings of the 30th ACM International Conference on Information and Knowledge Management (CIKM '21), November 1--5, 2021, Virtual Event, QLD, Australia}
\acmPrice{15.00}
\acmDOI{10.1145/3459637.3482360}
\acmISBN{978-1-4503-8446-9/21/11}

\begin{document}

\title[L$^2$NAS: Learning NAS via Continuous-Action RL]{L$^2$NAS: Learning to Optimize Neural Architectures via Continuous-Action Reinforcement Learning}

\author{Keith G. Mills$^{1\dagger}$, Fred X. Han$^{2}$, Mohammad Salameh$^2$, Seyed Saeed Changiz Rezaei$^2$}
\author{Linglong Kong$^1$, Wei Lu$^2$, Shuo Lian$^3$, Shangling Jui$^3$, Di Niu$^1$}
\thanks{$\dagger$Work done during an internship at Huawei Technologies Canada}
\affiliation{$^1$University of Alberta \city{Edmonton} \state{AB} \country{Canada}}
\affiliation{$^2$Huawei Technologies \city{Edmonton} \state{AB} \country{Canada}}
\affiliation{$^3$Huawei Kirin Solution \city{Shanghai} \country{China}}

\renewcommand{\shortauthors}{Mills et al.}

\begin{abstract}
\input{src/abstract}
\end{abstract}


\maketitle

\input{src/intro}
\input{src/method}
\input{src/spaces}

\input{src/results}
\input{src/related}
\input{src/conclusion}

\bibliographystyle{ACM-Reference-Format}
\balance
\bibliography{sample-base}


\end{document}

%% file: src/abstract.tex
Neural architecture search (NAS) has achieved remarkable results in deep neural network design.
Differentiable architecture search  converts the search over discrete architectures into a hyperparameter optimization problem which can be solved by gradient descent.
However, questions have been raised regarding the effectiveness and generalizability of gradient methods for solving non-convex architecture hyperparameter optimization problems. In this paper, we propose L$^{2}$NAS, which learns to intelligently optimize and update architecture hyperparameters via an actor neural network based on the distribution of high-performing architectures in the search history. We introduce a quantile-driven training procedure which efficiently trains L$^{2}$NAS in an actor-critic framework via continuous-action reinforcement learning. Experiments show that L$^{2}$NAS achieves state-of-the-art results on NAS-Bench-201 benchmark as well as DARTS search space and Once-for-All MobileNetV3 search space. We also show that search policies generated by L$^{2}$NAS are generalizable and transferable across different training datasets with minimal fine-tuning.

%% file: src/intro.tex
\section{Introduction}
\label{sec:intro}

Neural architecture search (NAS) automates neural network design and has achieved state-of-the-art performance in computer vision tasks\citep{liu2018DARTS}, \citep{tan2019efficientnet}, \citep{cai2020once}, \citep{li2020geometry}.  A NAS scheme usually consists of a search strategy and a performance estimation strategy for candidate architectures. Typical search strategies include random search \citep{li2019RS}, Bayesian optimization \citep{kandasamy2018neural}, and reinforcement learning \citep{pham2018ENAS, zoph2018learning}.

Differentiable architecture search (DARTS) \citep{liu2018DARTS} and the variants based on optimization~\citep{chen2019progressive, wu2019fbnet, xu2020pcdarts, chen2020stabilizing, li2020geometry, cai2020once} have dramatically reduced the search cost in NAS by converting search in an exponentially large space into optimization performed on a weight-sharing \textit{supernet}, where all candidate architectures are mixed through the architecture hyperparameters $\alpha$. The architecture hyperparameters $\alpha$ can then be optimized with gradient descent to minimize the meta loss of the supernet on a validation set.

However, differentiable architecture search has brought about a few issues. First, during search DARTS aims to minimize the validation loss of the mixed supernet, whereas the final evaluation is performed on a discrete architecture extracted from the supernet based on the largest architecture hyperparameters. Such a discrepancy leads to the optimization gap known as the discretization error~\citep{yu2019evaluating, xie2020weight} in NAS literature. Second, although stochastic gradient descent (SGD) has been remarkably effective for training neural network weights~\citep{wilson2017marginal}, it is unclear whether gradient descent extends to the bi-level optimization of both architecture hyperparameters and model weights, where the former can lie in non-Euclidean domains~\citep{li2020geometry}. Recent work shows that DARTS is unstable and unable to generalize to different search spaces, some of which contain dummy operators that DARTS fails to rule out during the search~\citep{zela2019understanding}.

Another common problem facing differentiable NAS approaches is a lack of exploration mechanisms. By design, gradient-based methods seek the nearest local loss minima as quickly as possible. Consequently, gradient-based NAS approaches are reported to be driven toward regions of the search space where the supernet can train rapidly. This results in wide and shallow architectures being selected over deeper and narrower networks~\citep{shu2019understanding}. The above issues have given motivation to developing more advanced schemes than gradient descent for differentiable neural architecture search.

In this paper, we propose L$^2$NAS which given a search space of candidate architectures, learns \textit{how} to optimize the architecture hyperparameters $\alpha$ in differentiable NAS, in order to achieve generalizable hyperparameter optimization policies in NAS that are transferable across training datasets.
Rather than always descending in the direction of gradients, L$^2$NAS updates $\alpha$ through a trainable actor neural network taking as input the statistical information learned from the search history. In other words, L$^2$NAS learns to learn (optimize) a search policy for $\alpha$.  In particular, we make the following contributions:

First, L$^2$NAS learns to generate architecture hyperparameters $\alpha$ in a continuous domain via an actor network based on MLP. The hyperparameters $\alpha$ then undergo a many-to-one deterministic mapping into a discrete architecture, based on which the reward of $\alpha$ can be obtained from either a weight-sharing supernet or from other performance estimation mechanisms such as querying true performance, if available. We design a novel state representation of the optimization landscape by performing average pooling of top $K$ architectures found in the search history, which will be fed into the actor to determine $\alpha$ in the next step.  Thus, L$^2$NAS can also be understood as replacing the gradient descent update rule with a meta-learned architecture hyperparameter optimizer represented by the actor.

Second, we propose an effective and efficient procedure to train L$^2$NAS  through a quantile-driven loss. Leveraging the Actor-Critic framework, we introduce a critic network to predict the reward given $\alpha$. Due to the stochastic nature of the reward model, the critic is specifically trained by the check loss in quantile regression \citep{koenker05Quantile} to predict the 90th or greater percentile of performance instead of expected mean performance under each $\alpha$ to track high potential regions in the search domain. The actor network is then updated through an efficient gradient ascent algorithm based on the critic output, following a similar framework in continuous-action reinforcement learning \citep{lillicrap2016DDPG, mills2021DDAS}.

Third, we incorporate several exploration strategies into L$^2$NAS, including a noise-based exploration mechanism and the $\epsilon$-greedy strategy, to ensure that the search space be adequately explored to avoid premature convergence.

Through extensive experiments, we show that L$^2$NAS achieves state-of-the-art performance on the public benchmark set NAS-Bench-201~\citep{dong2019bench} in terms of accuracy constrained by the number of architecture evaluations incurred, which converts to search cost in reality. In DARTS~\citep{liu2018DARTS} and Once-for-All (OFA)~\citep{cai2020once} search spaces, we show that L$^2$NAS can find models that achieve competitive accuracies on CIFAR-10 and ImageNet, as compared to existing methods operating in the same search spaces. Furthermore, we show that the architecture optimization policy pre-trained on a simpler dataset, e.g., CIFAR-10, can be transferred to search for competitive architectures that perform well on more complex or higher-resolution datasets, e.g., CIFAR-100 and ImageNet, after a small amount of fine-tuning, further reducing the cost of customizing architecture searches per dataset.

%% file: src/method.tex
\section{Proposed Method}
\label{sec:method}

Neural architecture search can be formulated as a black-box optimization problem \citep{chen2016learning, chen2017learning}.
Suppose that $x$ is a neural architecture. Let $S(x)$ be a real-valued function representing its performance after fully training the network on a given dataset, i.e., the accuracy of $x$. In NAS, the goal is to maximize $S(x)$ subject to $x\in \mathcal X$, where $\mathcal X$ denotes a predefined search space of neural architectures. The black-box function $S$ is not available in a closed form, but can be evaluated at any query point $x\in \mathcal X$.  In other words, $S$ can only be observed through point-wise observations.

Since the size of $\mathcal X$ grows exponentially in terms of the number of operators considered, differentiable architecture search \cite{liu2018DARTS} relaxes $\mathcal X$ into a continuous space of architecture representations, $\alpha$, a.k.a. \textit{architecture hyperparameters}, converting the search in $\mathcal X$ into an optimization problem over $\alpha$ in a continuous domain. That is, one can define a deterministic mapping function ${\tt Discretize}$ such that $x = {\tt Discretize}(\alpha)$, and maximizing $S(x)$ is converted to maximizing $S(\alpha)$.

\subsection{Continuous Relaxation of Discrete Architectures}
\label{sec:rlSpace}

We now describe how the continuous relaxation of a neural architecture can be achieved using DARTS \citep{liu2018DARTS} as an example. A neural network can be represented by a directed acyclic graph (DAG) $G = (N, E)$ with $|N|$ numbered nodes $1,2,\ldots, |N|$ and a set of directed edges $E$ connecting them. The nodes are latent data representations in a neural network. 
Let $x_i$ denote the latent representation on node $i$. 
Let $\mathcal{O}$ denote the predefined set of candidate operations.

In a discrete architecture in DARTS, each edge represents one operation in $\mathcal O$. 
But in its continuous relaxation $G$, a.k.a. \textit{the supernet}, each edge $(i,j)$ performs a weighted sum of all candidate operations applied onto $x_i$. 
Specifically, let $\alpha_{(i,j), o} \in \mathbb R$ be the raw weight value of operation $o$ on edge $(i,j)$, 
the computation performed on the edge $(i,j)$ from node $i$ to node $j$ ($i < j$) is defined as:
\[
    f_{i,j}(x_i) = \sum_{o \in \mathcal{O}}^{} \dfrac{\text{exp}(\alpha_{(i, j), o})}{\sum_{o' \in \mathcal{O}}^{} \text{exp}(\alpha_{(i, j), o'})}\cdot o(x_i).
\]
The result of edge $(i,j)$ will be aggregated with other incoming edges at node $j$, if any, to yield
\[
    x_j = \sum_{i:(i,j)\in E}f_{i,j}(x_i).
\]

We could then define $\alpha = \{\alpha_{(i,j),o}\} \in \mathbb{R}^{\abs{E} \times \mathcal{|O|}}$ as the hyperparameters matrix, which is a relaxed representation of all possible networks in the search space.
By optimizing $\alpha$ and supernet weights $w$ to increase $S(\alpha)$, i.e., validation accuracy of the supernet under $\alpha$, $\alpha$ would eventually converge to the region of better performing architectures.
DARTS achieves this by updating $\alpha$ and $w$ using gradient descent in a bi-level optimization setup: 
\begin{align}
    \min_{\alpha} \quad & S(\alpha) =  \mathcal{L}_{\text{val}}(w^*(\alpha), \alpha) \\
    \text{s.t.} \quad & w^*(\alpha) = \arg\min_w \mathcal{L}_{\text{train}}(w, \alpha),
\end{align}
where all supernet weights $w$ are fit to the training set given $\alpha$, while $\alpha$ is found by minimizing the validation loss (of the entire supernet) on a separate validation set.

Unlike DARTS, we set $S(\alpha) = S({\tt Discretize}(\alpha))$ in L$^2$NAS to evaluate $\alpha$ directly by the performance of the individual architecture derived from $\alpha$ via discretization.
Similarly, we train supernet weights $w$ in L$^2$NAS by randomly sampling $\alpha$ for every input batch of data and discretizing it into an architecture,
and only updating the corresponding supernet weights used in the sampled architecture. By doing so, L$^2$NAS avoids the generalization gap of DARTS caused by evaluating supernet performance during the search while evaluating individual architectures in the test.  

As will be demonstrated in experiments, L$^2$NAS can operate on many search spaces other than DARTS that allow continuous relaxation. Furthermore, L$^2$NAS can work regardless of how the performance $S(\alpha)$ is queried, whether from a predictor or a weight-sharing supernet. 

\subsection{Learning to Optimize Architecture Hyperparameters}
\label{sec:rlComp}

\begin{figure}[t]
	\centering
	\includegraphics[width=3.3in]{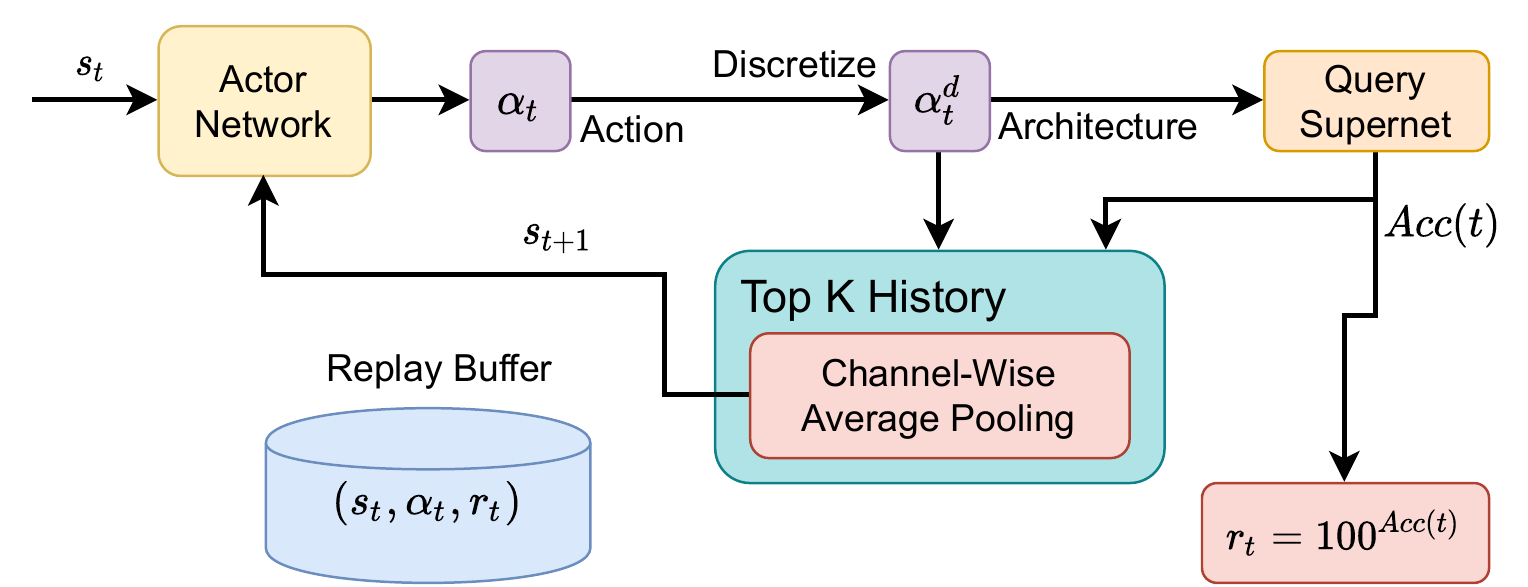}
	\caption{A High-Level Illustration of L$^2$NAS.}
	\label{fig:L2NASFlowChart}
	\vspace{-2mm}
\end{figure}

We now present the operating mechanisms of the proposed algorithm, L$^2$NAS. Traditionally, in differentiable architecture search, the architecture hyperparameters $\alpha$ are updated via gradient descent with a first-order or second-order approximation of the derivative of validation loss over $\alpha$. The goal of L$^2$NAS, however, is to learn an update rule $\mu$ through performing the following iterations: 
\begin{align}
    \alpha_t &= \mu(s_t),\nonumber\\
    r_t &\sim p(r|{\tt Discretize}(\alpha_t)),\nonumber\\
    s_{t+1} &= \phi (\alpha_t, r_t,s_t),\nonumber
\end{align}
where $s_t$ can be seen as a state recording important statistics in the optimization process up to time step $t$. At time step $t$, an action $\alpha_t$ is generated given the state $s_t$. We then discretize $\alpha_t$ into a neural architecture to query its performance $r_t$. Finally, the state $s_{t+1}$ is updated with the newly queried $\alpha_t$ and $r_t$. 
A high-level overview of our scheme is illustrated in Figure~\ref{fig:L2NASFlowChart}. 

Specifically, at step $t$, L$^2$NAS produces an action that serves as continuous architecture hyperparameters, i.e., $\alpha_t \in \mathbb{R}^{\abs{E} \times \mathcal{|O|}}$, through a neural network $\mu$, which is a multi-layer perceptron (MLP). Each continuous action $\alpha_t$ is deterministically mapped into a discrete $\alpha_t^d$. We further elaborate on this {\tt Discretize} process in Section~\ref{sec:spaces}, as it is search space dependent. Each $\alpha^d$ corresponds to an individual architecture in the search space. The reward of taking the query $\alpha_t$ is defined as
\begin{equation}
    \label{eq:reward}
    \centering
    r_t = 100^{Acc(\alpha_t^d)},
\end{equation}
where \textit{Acc} is the measured accuracy of the architecture $\alpha_t^d = {\tt Discretize}(\alpha_t)$.

Throughout the optimization process, L$^2$NAS keeps track of the top-$K$ architectures in terms of the accuracy seen so far. That is, we store a history tensor, $h_t  \in \{0, 1\}^{K \times \abs{E} \times \mathcal{|O|}}$ of top-$K$ $\alpha_t^d$ seen so far and define the state, $s_t \in \mathbb{R}^{\abs{E} \times \mathcal{|O|}}$, as the averaging of $h_t$ over the first dimension.
The state $s_t$ is meant to provide statistical information regarding the search space. 

Given the above definition, each entry of the state matrix $s_t$ represents the sample probability that a specific operation is present on an edge in the top-$K$ architectures seen so far. A higher value on an entry indicates that the corresponding operation-edge pair is favoured by top performing architectures. 

\subsection{Policy Training Procedure}
\label{sec:rlNet}

L$^2$NAS adopts an \textit{Actor-Critic} framework in reinforcement learning to train the update policy $\mu$. L$^2$NAS produces continuous actions, which makes its training procedure similar to the Deep Deterministic Policy Gradient (DDPG)~\citep{lillicrap2016DDPG} framework. In continuous-action RL, the agent interacts with an environment over an infinite number of steps, $t = 0, 1,...,$, by performing an action $a_t$ at each step, receiving a reward $r_t$ in return, followed by transitioning to a state $s_{t+1}$. Nevertheless, to solve our black-box optimization problem, specifically neural architecture search, there are several key differences in the policy training of L$^2$NAS from DDPG, including a redefined critic network, a quantile-driven loss in critic training, and the use of epsilon-greedy strategy to reduce the number of queries (architecture evaluations). 

In particular, we maintain two neural networks in L$^2$NAS: The actor, $\mu(s_t)$, which generates an action $\alpha_t$ given $s_t$; and the critic, $Q(\alpha_t)$, which predicts the action value. Note that as opposed to DDPG, we remove the dependency of the critic on $s_t$.
We also maintain a replay buffer, $R$, which stores triplets in the form of $(s_t, \alpha_t, r_t)$.
 
The actor network is given by,
\begin{equation}
    \centering
    \label{eq:actor}
    \alpha_t = \mu(s_t) + z_t,
\end{equation}
where $z_t = \tt Uniform (-\xi, \xi)$ is a small noise following a uniform distribution added to the actor output to encourage exploration. Furthermore, we introduce additional exploration strategies in the form of taking random actions drawn from a uniform distribution, $\tt Uniform(0, 1)^{\abs{E} \times \mathcal{|O|}}$, instead of being determined by the actor network in Equation~\ref{eq:actor}. We apply two different exploration strategies depending on the search space:
\begin{itemize}
    \item \textbf{$\epsilon$-greedy}: At every step, the actor will take a random action with probability $\epsilon$. We initialize $\epsilon$ to a high value and anneal it to a minimum value over time.
    \item \textbf{Random warm-up}: The agent takes random actions in the first $W$ steps. Actions taken during all remaining steps $t > W$ are determined by Equation~\ref{eq:actor}.
\end{itemize}

The original DDPG critic $Q(s_t, a_t)$ calculates a discounted estimate of future rewards~\citep{lillicrap2016DDPG} based on state transitions. In contrast, L$^2$NAS uses a critic $Q(\alpha_t)$ to directly approximate the reward $r_t$. It differs from DDPG in that the critic network does not take the state $s_t$ as an input. The critic network of L$^2$NAS is also an MLP with 3 hidden layers. 

Furthermore, considering the stochastic nature of $r_t$,
if the critic were trained by an $L_2$ loss as in DDPG, it would predict the conditional mean reward,  
\[r_t = \mathbb E[r|\alpha_t].\] 
In contrast, we use a \textit{check loss} to train the critic in L$^2$NAS. In quantile regression \citep{koenker05Quantile}, the $\tau$-th quantile of a random variable $Z$, denoted by $Q_{\tau}(Z)$, is defined as the value such that $Z$ is no more than $Q_{\tau}(Z)$ with probability $\tau$ and no less than $Q_{\tau}(Z)$ with probability $1-\tau$. When $\tau=0.5$, $Q_{0.5}(Z)$ is the median of $Z$. It is shown~\cite{koenker05Quantile} that 
\begin{eqnarray*}
Q_{\tau}(Z)=\arg\min_{c}\mathbb E_Z[\rho_{\tau}(Z-c)],
\end{eqnarray*}
where $\rho_{\tau}$ is the \emph{check loss} function defined as
\begin{eqnarray*}
  \rho_{\tau}(x)=
  x(\tau-\mathbf 1(x\leq 0)),
\end{eqnarray*}
where $\mathbf 1(x\leq 0)$ is one if $x\leq 0$ and zero otherwise, and $\tau \in [0, 1]$ corresponds to the desired quantile level.

L$^2$NAS is trained off-policy, which means that the experience $(s_t, \alpha_t, r_t)$ of a step $t$ is not immediately used to update the critic and actor networks. Instead, all experiences are stored in a replay buffer $R$ and randomly sampled in batches with replacement to train the actor and critic. At every time step $t$, we randomly sample a batch $B_R$ from the experience replay buffer and use it to update the critic network using the \textit{check loss}:
\begin{equation}
    \label{eq:checkCritic}
    \mathcal{L}_{\text{Critic}} = \frac{1}{\abs{B_R}} \sum_{i\in B_R}
    \rho_\tau(r_i - Q(\alpha_i))
\end{equation}
The actor network is then updated directly based on critic outputs with the following loss:
\begin{equation}
    \label{eq:actorUpdate}
    \centering
    \mathcal{L}_{\text{Actor}} = \frac{1}{\abs{B_R}} \sum_{i\in B_R}^{}Q(\mu(s_{i})).
\end{equation}

A critic trained by the check loss as in Equation~\ref{eq:checkCritic} learns to predict the $\tau$th conditional quantile of the reward 
\[
    r_t = Q_\tau(r|\alpha_t) = \arg\min_{c}\mathbb E_r[\rho_{\tau}(r-c)|\alpha_t].
\]
The key advantage is that the proposed critic is capable of picking up and dealing with the best architectures while the traditional $L_2$ loss is only able to handle average architectures. 
Put succinctly, the check loss function ensures that the critic predicts the tail of the reward, which corresponds to the accuracy of the best architectures selected. This knowledge is then used in actor update in Equation~\ref{eq:actorUpdate}.

Finally, we can accelerate policy training by updating the actor and critic networks with $C$ batches of samples pulled from the replay buffer per step, instead of one batch per step as in DDPG. This is useful in situations where the number of steps allowed is tightly budgeted. We determine the number of batches $C$ used per step by
$   
    C = \min(\floor*{\frac{\abs{R}}{\abs{B_R}}}, C_{\max}),
$
where $\abs{R}$ is the total number of samples in the replay buffer and $C_{\max}$ is an upperbound on $C$. The actor and critic networks start training once the experience replay buffer has accumulated $\abs{B_R}$ samples.

\subsection{Transferability}
\label{sec:transferability}

It is possible to train an agent on one dataset, e.g. CIFAR-10, and then transfer the pretrained agent to another dataset, e.g., CIFAR-100 or ImageNet, where both the critic and actor networks can be fine-tuned with a low number of steps in order to search for high-performance architectures based on the new dataset. 
Transferability of search policy can be achieved when the action space is held constant. 

For this purpose, we introduce another reward function such that the agent can generalize across different datasets and accuracy ranges. 
For example, the highest validation accuracies achieved by NAS-Bench-201~\citep{dong2019bench} on CIFAR-10 and CIFAR-100 are $94.37\%$ and $73.51\%$, respectively, which fall in different ranges. If a search algorithm is trained on CIFAR-10 and then transferred to CIFAR-100, it will need to be rescaled such that it learns a dataset-agnostic understanding on what constitutes a high-performance architecture. In L$^2$NAS, we accomplish this by rescaling the reward function to
\begin{equation}
    \centering
    \label{eq:transfer_reward}
    r_t = \dfrac{100^{\nicefrac{Acc(\alpha_t^d)}{Acc(Env)}}}{100} - 1,
\end{equation}
where $Acc(Env)$ is an environment-dependent accuracy measure used to scale the accuracy of architectures generated by the agent to fit to the target dataset.

%% file: src/spaces.tex
\section{Experimental Setup}
\label{sec:spaces}

In this section we enumerate our three candidate search spaces, NAS-Bench-201, DARTS and Once-for-All in detail. Specifically, we describe the internal layout, explain how discretization is performed and elaborate on our training details.

\subsection{NAS-Bench-201}
\label{sec:nb201Top}

The NAS-Bench-201 (NB) search space is based on NASNet~\cite{zoph2018learning}. Many cells are stacked repeatedly to form a neural network. There are two types of cells: \textit{normal cells}, which do not modify the dimensions of input tensors and \textit{reduction cells}, which are responsible for halving the height and width of input tensors while doubling the number of channels. The internal structure of reduction cells is fixed, and there are two in total, residing 1/3 and 2/3 through the network, respectively. All other cells are normal and the structure of normal cells is variable, facilitating architecture search.

The internal structure of the normal cells is represented by a DAG with an input node, an output node and $\abs{N}_{\text{NB}} = 3$ intermediate nodes. The input node of cell $k$ receives data from the output node of cell $k-1$. Each intermediate node is connected to the output node by an edge. Additionally, $\abs{E}_{\text{NB}} = 6$ edges connect the input and intermediate nodes. These edges are responsible for performing operations within the cell. The pre-defined operator set, $\mathcal{|O|}_{\text{NB}} = 5$, consists of the following: None (Zero), Skip Connection, Average Pooling 3$\times$3, {\tt Nor\_Conv\footnote{{\tt Nor\_Conv} refers to a sequence consisting of \textit{ReLU-Conv-BN} (batch normalization)~\cite{dong2019bench}.}} 1$\times$1 and {\tt Nor\_Conv} 3$\times$3. 

This layout gives rise to $\alpha_{\text{NB}} \in \mathbb{R}^{6 \times 5}$ for a total of 15,625 architectures. Algorithm~\ref{alg:discretizeNB} describes the discretization process for NAS-Bench-201. Essentially, the process consists of an \texttt{argmax} operation performed on each row of $\alpha_{\text{NB}}$ to select the operators.

\begin{algorithm}[t]
  \caption{{\tt Discretize()} in NAS-Bench-201/OFA} \label{alg:discretizeNB}
  \begin{algorithmic}[1]
    \STATE \textbf{Input:} $\alpha \in \mathbb{R}^{\abs{E} \times \abs{\mathcal O}}$ 
    \STATE \textbf{Output:} $\alpha^d \in \{0,1\}^{\abs{E} \times \abs{\mathcal O}}$
    \STATE $\alpha^d = 0^{\abs{E} \times \abs{\mathcal O}}$ 
    \FOR{$k = 0, 1,..,\abs{E}-1$}
        \STATE $A = \alpha[k, :]$ 
        \STATE $i_k = \arg\max_{i}A_i$
        \STATE $\alpha^d[k, i_k] = 1$
    \ENDFOR
    \STATE Return $\alpha^d$
  \end{algorithmic}
\end{algorithm}

We do not need to train networks from scratch to evaluate L$^2$NAS on NAS-Bench-201 as the accuracy of each architecture is provided in the form of an API and lookup table.

\subsection{DARTS}
\label{sec:dartsTop}

The DARTS (DA) topology is also based on NASNet~\cite{zoph2018learning}, but more complex than NAS-Bench-201 and contains approximately 10$^{18}$~\citep{siems2020bench} architectures. Rather than simply receiving input from the previous cell, each cell $k$ receives input from the previous two cells, $k-1$ and $k-2$, respectively. There are $\abs{N}_{\text{DA}} = 4$ intermediate nodes forming $\abs{E}_{\text{DA}} = 14$ edges as each intermediate node can receive data from all previous intermediate nodes and both input nodes. There are $\mathcal{|O|}_{\text{DA}} = 7$ operations consisting of the following: Maximum Pooling 3$\times$3, Average Pooling 3$\times$3, Skip Connection, Separable Convolution 3$\times$3, Separable Convolution 5$\times$5, Dilation Convolution 3$\times$3 and Dilation Convolution 5$\times$5. Note that compared to the original DARTS, we omit the `None' operation. Although DARTS allows it during search, unlike NAS-Bench-201, it cannot be selected to form a discrete architecture once search is complete.

Unlike NAS-Bench-201, the reduction cell structure is not fixed and is searchable independently of the normal cell architecture. These properties give rise to architectures represented by two matrices, $\alpha_{\text{DA}}^{\text{N}} \in \mathbb{R}^{|E| \times \mathcal{|O|}} = \mathbb{R}^{14 \times 7}$ and $\alpha_{\text{DA}}^{\text{R}} \in \mathbb{R}^{|E| \times \mathcal{|O|}} = \mathbb{R}^{14 \times 7}$, corresponding to the normal and reduction cells, respectively. 

Discretization for DARTS is more complex than it is for NAS-Bench-201. Similar to~\cite{mills2021DDAS}, each intermediate node in $N$ will receive input from only 2 of the directed edges that feed into it. Each of these selected edges shall perform a single operation. The choice of which edges are chosen, and which operators will occupy these edges are determined using the magnitude of their architecture distribution parameters, where higher is better. This means that while the cell may contain up to $\abs{E} = \sum_{i=1}^{\abs{N}}(i + 1)$ edges during search, only $2|N|$ edges are allowed after discretization. 
Algorithm~\ref{alg:discretize} describes the process of discretizing either $\alpha_{\text{DA}}$ matrix.

\begin{algorithm}[t]
  \caption{{\tt Discretize()} in DARTS/PC-DARTS} \label{alg:discretize}
  \begin{algorithmic}
    \STATE \textbf{Input:} $\alpha \in \mathbb{R}^{\abs{E} \times \abs{\mathcal O}}$ 
    \STATE \textbf{Output:} $\alpha^d \in \{0,1\}^{\abs{E} \times \abs{\mathcal O}}$
    \STATE $\mbox{Start} = 0, n = 1$
    \STATE $\alpha^d = 0$ 
    \FOR{$k = 0, 1,..,\abs{N}-1$} 
        \STATE $\mbox{End} = \mbox{Start} + n$ 
        \STATE $A = \alpha[\mbox{Start}:\mbox{End}, :]$ 
        \STATE $(i_1, j_1) = \arg\max_{(i,j)}A_{ij}$
        \STATE $(i_2, j_2) = \arg\max_{(i,j):i\neq i_1}A_{ij}$ 
        \STATE $\alpha^d[\mbox{Start}+i_1, j_1] = 1$
        \STATE $\alpha^d[\mbox{Start}+i_2, j_2] = 1$
        \STATE $\mbox{Start} =\mbox{End} + 1$
        \STATE $n = n + 1$
    \ENDFOR
    \STATE Return $\alpha^d$
  \end{algorithmic}
\end{algorithm}

We train a weight-sharing supernet using the PC-DARTS framework, which uses the same topology and operation set as DARTS, but incorporates several memory-saving features~\citep{xu2020pcdarts}. We refer the reader to PC-DARTS for more details, but note that our use of Algorithm~\ref{alg:discretize} 
allows us to omit the use of `$\beta$'.

Our supernet training strategy is inspired by random search~\citep{li2019RS} and the single-path uniform sampling scheme used by \cite{guo2019Supernet}. DARTS supernets are only used for search; evaluation is performed by training found architectures from scratch. Moreover, this sampling method saves GPU memory cost \citep{dong2019GDAS, cai2018proxylessnas}.

For each batch of training data, we first sample a random matrix $\alpha \in \mathbb{R}^{\abs{E} \times \mathcal{|O|}}$, which is then discretized. The corresponding weights of the discrete architecture are then updated with the batch of data. This process is equivalent to selecting random individual architectures from the supernet to update per each batch of data.

We train a supernet for CIFAR-10. CIFAR-10 consists of 50k training samples and 10k test samples split across 10 classes. We split the original training set in half into equally sized training and validation partitions. The new training set is used to train the supernet. The validation set is used to query the supernet during model search. The supernet consists of 8 cells (6 normal, 2 reduction) with an initial channel multiplier of 16. The initial learning rate is set to 0.025 and is annealed down to $1e^{-3}$ by cosine schedule. 
Stochastic gradient descent with a momentum factor of 0.9 optimizes the weights. We use Cutout~\citep{devries2017cutout} using the recommended length for CIFAR-10 and train for 10k epochs with a batch size of 250. On a single RTX 2080 Ti, supernet training takes around 3 GPU days.

At the end of each epoch, we gauge the validation set accuracy of the supernet. We record the highest validation accuracy observed during supernet training. This allows the supernet to be used in conjunction with Equation~\ref{eq:transfer_reward}.

To facilitate transferability, we train additional supernets on CIFAR-100 and ImageNet-32-120\footnote{First 120 classes of ImageNet downsampled to 32x32 images.}~\citep{chrabaszcz2017downsampled} as it has been shown that ImageNet subsets can be used as efficient proxies~\cite{wu2019fbnet}. CIFAR-100 has the same number of training and test samples as CIFAR-10, only split across 100 classes instead of 10. Likewise, the supernet training scheme for CIFAR-100 follows that of CIFAR-10. ImageNet-32-120 consists of 155k training and 6k test samples split across 120 classes, respectively. We further split the training set into separate training and validation sets like the CIFAR datasets. The supernet trains for 5k steps using a batch size of 750 with the same optimizer and learning rate scheduler as the CIFAR datasets. 

After any search on DARTS is complete, we select the best architecture according to validation accuracy, and evaluate it by training it from scratch and obtaining its accuracy on the test set. For CIFAR datasets, DARTS evaluation is performed using models with 20 cells, 18 of which are normal while the remaining 2 are reduction. We set the initial channel size to 36. We utilize Cutout~\citep{devries2017cutout} with the recommended lengths and an auxiliary head with a weight of 0.4. The initial learning rate is 0.025, which is annealed down to 0 following a cosine schedule~\citep{Loshchilov2017SGDRSG} over 1,000 epochs. Batch size is 96 and drop path probability is 0.2.

ImageNet evaluation on DARTS is performed using the same hyperparameters as PC-DARTS~\cite{xu2020pcdarts}. The network consists of 14 cells (12 normal, 2 reduction) and is trained for 250 epochs. An initial learning rate of 0.5 is used and is annealed down to zero after an initial 5 epochs of warmup.

\begin{table*}[t]
	\centering
	\caption{Accuracies obtained on NAS-Bench-201 datasets compared to other methods. The horizontal line demarcates weight-sharing algorithms from those that directly query oracle information. We run L$^2$NAS for 500 and 1,000 steps per experiment across 10 different random seeds, and report the mean and standard deviation.}
	\label{table:nb201}
	\scalebox{0.9}{
	\begin{tabular}{l|cc|cc|cc}
		\toprule
		& \multicolumn{2}{c|}{\textbf{CIFAR-10}} &
		\multicolumn{2}{c|}{\textbf{CIFAR-100}} &
		\multicolumn{2}{c}{\textbf{ImageNet-16-120}} \\ \midrule
		\textbf{Method} & \textbf{Valid [\%]}  & \textbf{Test [\%]} & \textbf{Valid [\%]}  & \textbf{Test [\%]} & \textbf{Valid [\%]}  & \textbf{Test [\%]} \\ \midrule
		DARTS~\citep{liu2018DARTS} & 39.77 $\pm$ 0.00 & 54.30 $\pm$ 0.00 & 15.03 $\pm$ 0.00 & 15.61 $\pm$ 0.00 & 16.43 $\pm$ 0.00 & 16.32 $\pm$ 0.00 \\
		ENAS~\citep{pham2018ENAS} & 37.51 $\pm$ 3.19 & 53.89 $\pm$ 0.58 & 13.37 $\pm$ 2.35 & 13.96 $\pm$ 2.33 & 15.06 $\pm$ 1.95 & 14.84 $\pm$ 2.10 \\
		GDAS~\citep{dong2019GDAS} & 89.89 $\pm$ 0.08 & 93.61 $\pm$ 0.09 & 71.34 $\pm$ 0.04 & 70.70 $\pm$ 0.30 & 41.59 $\pm$ 1.33 & 41.71 $\pm$ 0.98 \\
		GAEA~\citep{li2020geometry} & -- & 94.10 $\pm$ 0.29 & -- & \textbf{73.43} $\pm$ \textbf{0.13} & -- & 46.36 $\pm$ 0.00 \\ 
		\midrule
		RS~\citep{dong2019bench} & 90.93 $\pm$ 0.36 & 93.70 $\pm$ 0.36 & 70.93 $\pm$ 1.09 & 71.04 $\pm$ 1.07 & 44.45 $\pm$ 1.10 & 44.57 $\pm$ 1.25 \\
		REA~\citep{dong2019bench} & 91.19 $\pm$ 0.31 & 93.92 $\pm$ 0.30 & 71.81 $\pm$ 1.12 & 71.84 $\pm$ 0.99  & 45.15 $\pm$ 0.89 & 45.54 $\pm$ 1.03 \\
		REINFORCE~\citep{dong2019bench} & 91.09 $\pm$ 0.37 & 93.85 $\pm$ 0.37 & 71.61 $\pm$ 1.12 & 71.71 $\pm$ 1.09 & 45.05 $\pm$ 1.02 & 45.24 $\pm$ 1.18 \\
		BOHB~\citep{dong2019bench} & 90.82 $\pm$ 0.53 & 93.61 $\pm$ 0.52 & 70.74 $\pm$ 1.29 & 70.85 $\pm$ 1.28 & 44.26 $\pm$ 1.36 & 44.42 $\pm$ 1.49 \\
		\textit{arch2vec}-RL~\citep{yan2020does} & 91.32 $\pm$ 0.42 & 94.12 $\pm$ 0.42 & 73.12 $\pm$ 0.72 & 73.15 $\pm$ 0.78 & 46.22 $\pm$ 0.30 & 46.16 $\pm$ 0.38 \\
		\textit{arch2vec}-BO & 91.41 $\pm$ 0.22 & 94.18 $\pm$ 0.24 & \textbf{73.35} $\pm$ \textbf{0.32} & 73.37 $\pm$ 0.30 & 46.34 $\pm$ 0.18 & 46.27 $\pm$ 0.37 \\
		\textbf{L$^2$NAS-500} & 91.36 $\pm$ 0.19 & 94.11 $\pm$ 0.16 & 72.47 $\pm$ 0.74 & 72.69 $\pm$ 0.58 & 46.23 $\pm$ 0.28 & 46.74 $\pm$ 0.39 \\
		\textbf{L$^2$NAS-1k} & \textbf{91.47} $\pm$ \textbf{0.15} & \textbf{94.28} $\pm$ \textbf{0.08} & 73.02 $\pm$ 0.52 & 73.09 $\pm$ 0.35 & \textbf{46.58} $\pm$ \textbf{0.08} & \textbf{47.03} $\pm$ \textbf{0.27} \\ \midrule
		\textit{True Optimal} & \textit{91.61} & \textit{94.37} & \textit{73.49} & \textit{73.51} & \textit{46.77} & \textit{47.31} \\
		\bottomrule
	\end{tabular}
	}
\end{table*}

\subsection{Once-for-All}
\label{sec:ofaTop}

OFA, as originally proposed, uses MobileNetV3 (MBv3)~\citep{howard2019searching} as a backbone structure and provides pre-trained supernets with elastic depth and width, allowing architectures to be searched for in the MBv3 design space and evaluated on ImageNet~\cite{deng2009imagenet}. 

The search space contains approximately $10^{19}$ architectures. The supernet contains 10--20 layers, divided into 5 units, each of which contains 2--4 layers to give   $  \abs{E}_{\text{OFA}} = 20$. Each layer contains a block operation. The operation space consists of \textit{MBConv}~\citep{howard2019searching} blocks of the form {\tt MBConv-e-k} where `{\tt e}' refers to a channel expansion factor in $\{3, 4, 6\}$, and `{\tt k}' refers to a square kernel size in $\{3, 5, 7\}$, giving rise to $\mathcal{|O|}_{\text{OFA}} = 9$ different blocks.

\begin{figure}
	\centering
	\includegraphics[width=3.2in]{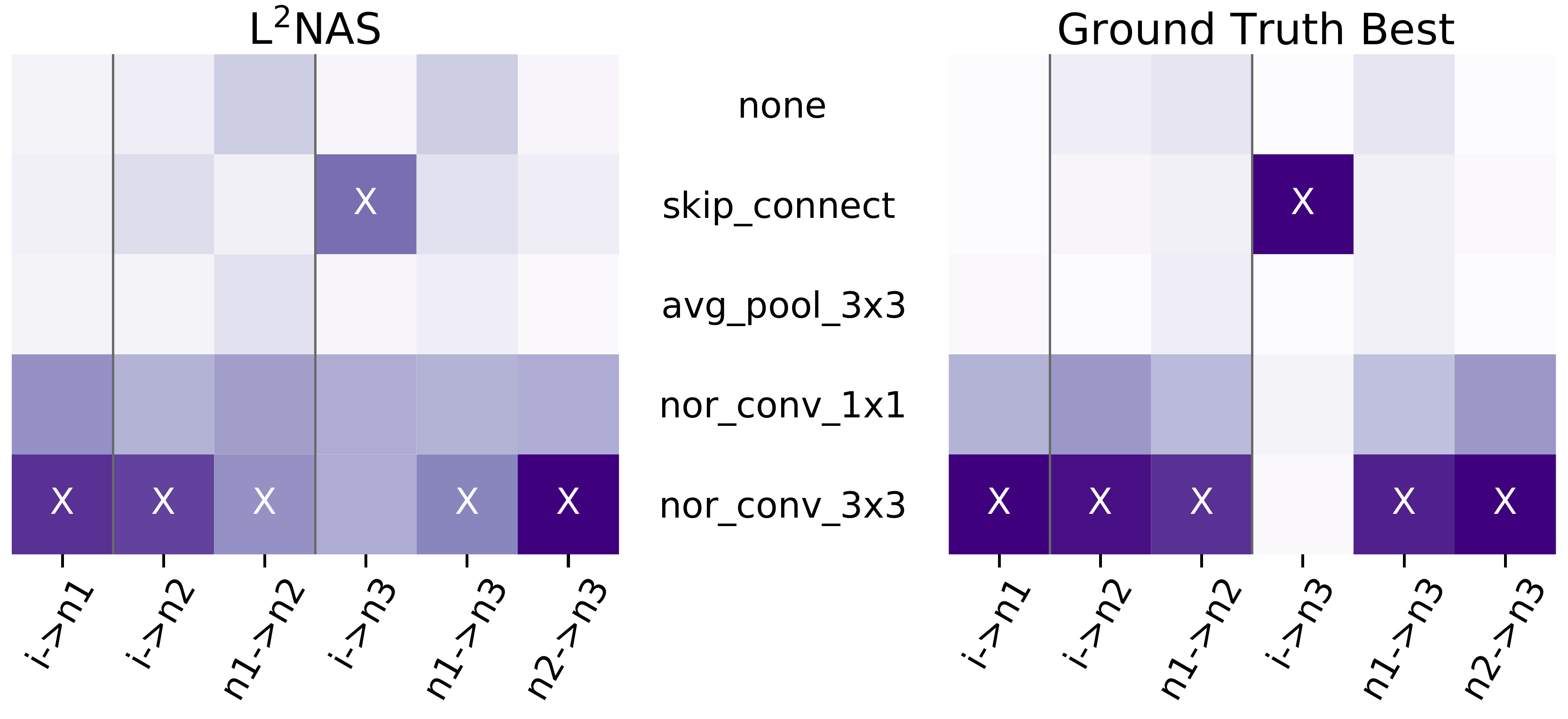}
	\caption{Comparison of the final state representation at the end of an L$^2$NAS search on NAS-Bench-201 CIFAR-100 test accuracy (left) with the average of $\alpha_d$ of the true top-$K$ architectures (right). $K = 64$. Rows indicate operations, columns indicate edges. `i' means input, `n' denotes an intermediate node. Darker color indicates higher values. Vertical bars demarcate destination nodes. `X' marks the operation-edge pairs corresponding to the best architecture.}
	\label{fig:groundTruth}
	\vspace{-3mm}
\end{figure}

Therefore, for OFA, $\alpha_{\text{OFA}}$ consists of two matrices: $\alpha_{\text{OFA}, \mathcal{O}} \in \mathbb{R}^{20 \times 9}$, which controls the operations selected for each layer and $\alpha_{\text{OFA}, D} \in \mathbb{R}^{5 \times 3}$, which controls the depth of the network. Discretization is performed using Algorithm~\ref{alg:discretizeNB} separately on each matrix.

The publicly available OFA code repository\footnote{https://github.com/mit-han-lab/once-for-all} provides pre-trained supernets and an API for evaluating individual architectures on the ImageNet validation set, which contains 50k 224$\times$224 color images spread across 1000 classes. We make use of these resources. In particular, we use both versions of the MBv3 supernets, which we denote as \textit{OFA}\footnote{{\tt ofa\_mbv3\_d234\_e346\_k357\_w1.0}} and \textit{OFA$_{Large}$}\footnote{{\tt ofa\_mbv3\_d234\_e346\_k357\_w1.2}}.

The only difference between these two supernets is the base channel width multiplier, which is not searchable. By using both of these supernets we are able to test and compare L$^2$NAS across more than 1 version of the search space.

\subsection{DDPG Agent}
\label{sec:ddpgAgent}

Our DDPG agent consists of 2 neural networks: an actor and a critic. Both are multi-layer perceptrons with 3 hidden layers. The actor produces vectorized actions $a_t$ which are then split if $\alpha$ consists of more than one matrix (e.g., DARTS and OFA) and then re-shaped into matrices. The number of neurons in each hidden layer is 128 for NAS-Bench-201 and 256 for DARTS and OFA as their respective $\alpha$ matrices are larger. ReLU~\citep{hinton10Relu} serves as the activation function in the hidden layers while the final layer of the actor network features a sigmoid activation. The critic uses an identity function as it produces a scalar. Both networks are optimized using Adam~\citep{kingma14Adam} with $\vec{\beta} = (0.9, 0.99)$. Learning rates for the actor and critic networks are $1e^{-8}$ and $1e^{-4}$, respectively.

%% file: src/results.tex
\section{Results}
\label{sec:results}

In this section, we first evaluate L$^2$NAS on NAS-Bench-201~\citep{dong2019bench}. We then perform search experiments with L$^2$NAS in the DARTS search space based on CIFAR-10 and compare against other state-of-the-art methods that also operate in the same search space. We also present search results in the OFA search space evaluated on ImageNet. Finally, we show the feasibility of transferring a search policy found by L$^2$NAS on a smaller CIFAR-10 dataset to architecture searches performed on CIFAR-100 and ImageNet with superior efficiency.

\subsection{NAS Benchmark Performance}
\label{sec:nb201}

NAS-Bench-201 architectures are evaluated on three datasets: CIFAR-10, CIFAR-100 and ImageNet-16-120\footnote{First 120 classes of ImageNet downsampled to 16$\times$16 images.}~\citep{chrabaszcz2017downsampled}. Therefore, in this context, our goal is to use the accuracy information provided by NAS-Bench-201 to find the best performing architecture within a small number of queries (steps) to the benchmark, since each query for accuracy converts into an architecture evaluation in reality.

We set $K = 64$, $\tau = 0.9$, $|B_R| = 8$, $\xi = 1e^{-4}$
and $C_{max} = 10$. On NAS-Bench-201, L$^2$NAS performs exploration using $\epsilon$-greedy. The initial value of $\epsilon$ is $1.0$ and it is annealed via a cosine schedule to a minimum of $0.05$ by step $t = 175$. We use Equation~\ref{eq:reward} to calculate NAS-Bench-201 rewards. 1000 steps of L$^2$NAS on NAS-Bench-201 executes in less than 4 minutes when only using a CPU.

Table~\ref{table:nb201} shows the results in comparison with a range of popular search methods with NAS-Bench-201 results reported. The methods in the first category, i.e., DARTS, ENAS, GDAS, GAEA, are gradient-based and must operate on a weight-sharing supernet, while the other methods like L$^2$NAS can simply perform search by querying the ground-truth oracle performance. We observe that with only 500 steps (queries), L$^2$NAS achieves high performance and with 1000 steps, achieves state-of-the-art performance on CIFAR-10 and ImageNet-16-120. Indeed, it is worth noting that both L$^2$NAS-1k and L$^2$NAS-500 clearly outperform all other methods in the search on test set of ImageNet-16-120. The ability of L$^2$NAS to find top architectures within a small number of queries stems from the effective strategy to balance exploration and exploitation.

Note that \textit{arch2vec}, achieving better results on CIFAR-100 validation set, pre-trains architecture embeddings using unsupervised learning before the search. Therefore, in fact, L$^2$NAS can be complemented by the architecture representation learning in \textit{arch2vec} to further enhance search efficiency. GAEA results are from searches performed on a weight-sharing supernet and is thus not comparable to results in the second category. 

Figure~\ref{fig:groundTruth} shows the final state representation $s_{t=500}$ of a 500-step L$^2$NAS experiment as compared to the ground-truth best architectures in NAS-Bench-201. We observe that the average $\alpha^d$ of the top-$K$ ($K=64$) architectures found by L$^2$NAS after querying only a fraction of NAS-Bench-201 is very close to that of the true top-$K$ architectures. This implies that L$^2$NAS can effectively and efficiently charter a search space.

\subsection{CIFAR-10 Performance on DARTS}
\label{sec:dartsEvaluation}

Next, we evaluate the performance of L$^2$NAS searching in the DARTS search space based on CIFAR-10 using weight-sharing and compare with other state-of-the-art results. Given that DARTS dwarfs NAS-Bench-201 by many magnitudes~\cite{siems2020bench} and the ground truth accuracy of every architecture is not known, we do not focus on finding the best architecture in the least number of queries. Instead, we adjust our search to perform additional emphasis on exploration so a more thorough sweep of the search space is performed while the critic focuses on learning a narrower accuracy quantile.

We set $K = 500$, $\tau = 0.95$, $\abs{B_R} = 64$, $\xi = 5e^{-5}$ and $C_{max} = 1$. We run the agent for 20k steps and limit the replay buffer to contain only the last 5k experiences. Exploration is achieved using random warm-up with $ W = 3000 $. Like NAS-Bench-201, we use Equation~\ref{eq:reward} to calculate the reward. Architecture search takes roughly 1 GPU day on a single RTX 2080 Ti. 

\begin{table}[t]
	\centering
	\caption{CIFAR-10 Results. Methods in the second and third categories search on P-DARTS and DARTS search spaces, respectively.}
	\label{table:cifar10Result}
	\scalebox{0.9}{
	\begin{tabular}{l|cc}
		\toprule
		\textbf{Architecture}  & \textbf{Test Acc. [\%]}&  \textbf{Params [M]} \\ \midrule
        ENAS~\citep{pham2018ENAS} & 97.11 & 4.6 \\
        GDAS~\citep{dong2019GDAS} & 97.18 & 2.5 \\
        AlphaX~\citep{wang2018neural} & 97.46 $\pm$ 0.06 & 2.8 \\ \midrule
        P-DARTS~\citep{chen2019progressive} & 97.50 & 3.4 \\
        P-SDARTS~\citep{chen2020stabilizing} & 97.52 $\pm$ 0.02 & 3.4 \\ \midrule
        DARTS 1st~\citep{liu2018DARTS} & 97.00 $\pm$ 0.14 & 3.3 \\
        DARTS 2nd & 97.24 $\pm$ 0.09 & 3.3 \\
        SNAS~\citep{xie2018SNAS} & 97.15 $\pm$ 0.02 & 2.8 \\ 
        EcoNAS~\citep{zhou2020econas} & 97.38 $\pm$ 0.02 & 2.9 \\
        ISTA-NAS 2S~\citep{yang2020ista} & 97.46 $\pm$ 0.05 & 3.3 \\
        \textit{arch2vec}-RL~\citep{yan2020does} & 97.35 $\pm$ 0.05 & 3.3 \\
        \textit{arch2vec}-BO & 97.44 $\pm$ 0.05 & 3.6 \\
        MdeNAS~\citep{zheng2019multinomial} & 97.45 & 3.6 \\
        MiLeNAS~\citep{he2020milenas} & 97.49 $\pm$ 0.04 & 3.9 \\
        PC-DARTS~\citep{xu2020pcdarts} & 97.43 $\pm$ 0.07 & 3.6 \\
        PC-SDARTS & 97.51 $\pm$ 0.04 & 3.5 \\
        PC-GAEA~\citep{li2020geometry} & 97.50 $\pm$ 0.06 & 3.7 \\
        \textbf{L$^2$NAS} & \textbf{97.51} $\pm$ \textbf{0.12} & \textbf{3.8} \\
		\bottomrule
	\end{tabular}
	}
	\vspace{-2mm}
\end{table}

\begin{figure*}
    \centering
    \subfloat[L$^2$NAS OFA Architecture]{
        \includegraphics[angle=90,width=6.2in]{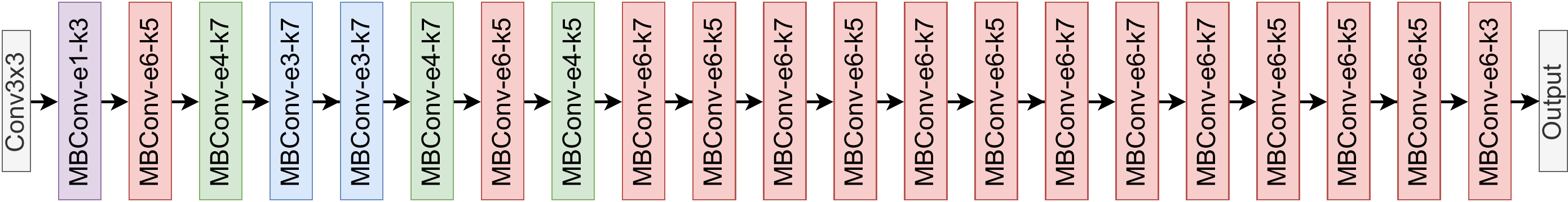}}
    \\
    \vspace{-3mm}
    \subfloat[L$^2$NAS OFA$_{Large}$ Architecture]{
        \includegraphics[angle=90,width=6.2in]{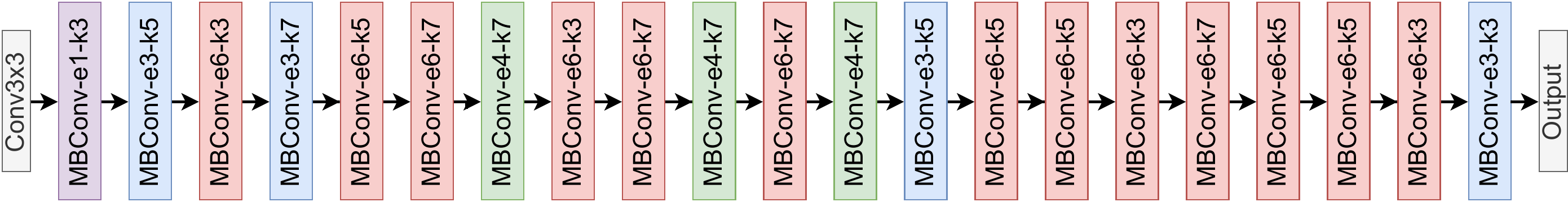}}
    \vspace{-3mm}
    \caption{Architectures found on the OFA (top) and OFA$_{Large}$ (bottom) supernets; initial {\tt MBConv-e1-k3} block is not searchable.}
    \label{fig:ofa_nets}
\end{figure*}

Table~\ref{table:cifar10Result} lists the results as compared to a wide range of other algorithms. 
For fair comparisons, we specifically compare to NAS methods that also reported performance exactly on the DARTS search space, as shown in the third category of Table~\ref{table:cifar10Result}, while the first category illustrates cell search methods using other search spaces.
We measure test accuracies over 5 evaluation runs for the best architecture found by L$^2$NAS.
Other results are the top results taken from their respective publications.

DARTS achieved 97.24\%, while PC-DARTS increased the accuracy to 97.43\% before GAEA (PC-GAEA) further increased it to 97.50\% on average with a maximum accuracy of 97.61\% among independent runs. The architecture found by L$^2$NAS attained an average accuracy of 97.51\%, with a maximum of \textbf{97.64\%}, which exceeds the performance of other state-of-the-art results achieved by GAEA and S-DARTS. At 3.8M parameters, our architecture is also more efficient than other larger ones like MiLeNAS.

It is worth noting that schemes based on Progressive DARTS (P-DARTS)~\citep{chen2019progressive} are listed separately from other methods that exactly operate on DARTS search space. P-DARTS uses multiple supernets and gradually prune the number of operations are while the network depth is increased during search. Additionally, restrictions are placed on the number of times specific operations can be selected. Since we do not employ these features, they are not a fair comparison. Furthermore, it is also worth noting that the `One-stage' (1S) variant of ISTA-NAS  achieved an even higher average accuracy. However, it
employs unique and novel modifications to the original operation set of DARTS which boosts the performance. Thus, it is not listed for fair comparison. In contrast, the `Two-stage' (2S) version of ISTA-NAS operates in the same DARTS search space and is thus listed here. Finally, PC-DARTS enables partial channel connections to enhance the memory efficiency of DARTS during search. It still uses the DARTS supernet without altering operations, and therefore is a fair comparison.

\subsection{ImageNet Performance}
\label{sec:imageNetEval}

\begin{table}[b]
	\centering
	\caption{Comparison of L$^2$NAS with other state-of-the-art architectures on ImageNet.} 
	\label{table:imagenetResult}
	\scalebox{0.9}{
	\begin{tabular}{l|ccc}
		\toprule
		\textbf{Architecture}  & \textbf{Top-1 Acc. [\%]}  & \textbf{Top-5 Acc. [\%]}&  \textbf{MACs [M]}\\ \midrule
		PC-SDARTS & 75.7 & 92.6 & -- \\
		P-SDARTS & 75.8 & 92.8 & -- \\
		PC-GAEA & 76.0 & 92.7 & -- \\ \midrule
		MBv2 & 74.7 & -- & 585 \\
		ProxylessNAS & 75.1 & 92.5 & 320 \\
		MBv3-L 0.75 & 73.3 & -- & 155 \\
		MBv3-L 1.0 & 75.2 & -- & 219 \\ \midrule
		EfficientNet-B0 & 77.1 & 93.9 & 390 \\
		EfficientNet-B1 & 79.1 & 94.4 & 700 \\
		EfficientNet-B2 & 80.1 & 94.9 & 1000 \\ \midrule
		Cream-S & 77.6 & 93.3 & 287 \\
		Cream-M & 79.2 & 94.2 & 481 \\
		Cream-L & 80.0 & 94.7 & 604 \\ \midrule
		OFA & 76.0 & -- & 230 \\
		OFA$_{Large}$ & 79.0 & 94.5 & 595 \\
        \textbf{L$^2$NAS} & \textbf{77.4} & \textbf{93.4} & \textbf{467} \\
        \textbf{L$^2$NAS$_{Large}$} & \textbf{79.3} & \textbf{94.6} & \textbf{618} \\
		\bottomrule
	\end{tabular}
	}
\end{table}

We evaluate the ImageNet performance of L$^2$NAS by applying it to the search space of Once-for-All (OFA)~\citep{cai2020once} on MobileNetV3.

We use the same values for $K$, $\tau$, $\abs{B_R}$, $C_{max}$ and $W$ as DARTS. However, since ImageNet inference is costlier than CIFAR, we only train the agent for 10k steps with $\xi = 1e^{-4}$ for additional exploration. We use the rescaled reward Equation~\ref{eq:transfer_reward} with $Acc(Env)$ set to $76\%$ for OFA and $79\%$ for OFA$_{Large}$. We use L$^2$NAS to search for the architecture with the best accuracy on OFA supernet and report test accuracy by directly using model weights inherited from the supernet without further fine-tuning the weights. Search takes about 150 GPU hours or under 1 week on a single RTX 2080 Ti.

Figure~\ref{fig:ofa_nets} illustrates the OFA architectures found. Like NAS-Bench-201, L$^2$NAS learns to select large architectures in order to achieve high accuracy on OFA. Specifically, L$^2$NAS focuses on selecting the largest expansion ratio (6) in the latter half of the network.

ImageNet results are provided in Table~\ref{table:imagenetResult}, in which the last category lists methods that also search in OFA, while the other 4 categories list results of other methods obtained in other related search spaces involving different backbone setups and operation sets. 
These other related search spaces include the predecessor to MBv3, MobileNetV2 (MBv2)~\citep{sandler2018mobilenetv2}, which ProxylessNAS~\citep{cai2018proxylessnas} is based on, using different set of operations. Variants of MBv3 include EfficientNet~\citep{tan2019efficientnet}, which focuses on striking a balance between network depth and resolution. Cream of the Crop also uses MBv3 as the backbone structure, yet with a modified set of searchable operations introduced to boost performance.

From a comparison against the models OFA and OFA$_{Large}$, which are obtained through an evolutionary algorithm applied on the  supernet, we observe that L$^2$NAS outperforms both of these in terms of both top-1 accuracy, while both are directly using weights inherited from the OFA supernet for evaluation. Note that we do not compare to the result of OFA$_{Large}$ after fine-tuning it, which is also reported in \citep{cai2020once}. These findings further demonstrate the effectiveness of L$^2$NAS as a search method, excluding the factors of search space and evaluation methodology.

In addition, we are able to achieve a much higher top-1 and top-5 accuracy than any state-of-the-art DARTS architecture.
Additionally, we outperform EfficientNet-B0 and B1, while EfficientNet-B2 is a large architecture with more than 1 billion MACs.
Cream of the Crop, although a state-of-the-art method in the mobile regime, is not directly comparable to our result, due to the different search space and evaluation method used. It searches in a variant space of MBv3 with different operations than OFA, and trains its supernet for 120 epochs and then retrains specific architectures for 500 epochs. 

\subsection{Transferability of Search Policies}
\label{sec:transfer_eval}

Finally, we perform transferability tests by training a policy on a simpler dataset like CIFAR-10, and then perform policy fine-tuning on more complex datasets, including CIFAR-100 and ImageNet. This is in contrast to previous NAS methods like DARTS that search for an architecture on CIFAR-10, then directly transfer the architecture to ImageNet for evaluation. Other methods like PC-DARTS and GAEA perform direct searches on ImageNet, i.e., a separate search for each individual dataset must be done.

We use the rescaled reward in Equation~\ref{eq:transfer_reward} with $\tau = 0.95$ to pre-train a transferable policy on CIFAR-10. $Acc(Env)$ is set to the maximum accuracy observed during supernet training. Table~\ref{tab:os_acc} lists the highest accuracy for each supernet. We train the agent for 10k steps. The first $W = 3000$ steps are used for exploration and we set $\xi = 1e^{-4}$ to further increase exploration. The pretrained actor and critic networks are then fine-tuned on CIFAR-100 and ImageNet-32-120 supernets trained according to the procedure in Section~\ref{sec:dartsTop}.
We set $K = 100$ and fine-tune for a total of 1k steps, using the first $W = 500$ steps for exploration. Initial policy pretraining takes 12 GPU hours. We perform the fine-tuning task 5 times with different random seeds. Each run of fine-tuning takes about 1.5 and 2 GPU hours for CIFAR-100 and ImageNet, respectively.

Table~\ref{table:transferability} provides evaluation results. For comparison, we also evaluate the original architectures published by DARTS on CIFAR-100. For both CIFAR-100 and ImageNet, transferred policies can find architectures that actually outperform the ones found via direct search, as the search policies instead of architectures are migrated from CIFAR-10 to operate with more complex datasets. Moreover, the cost of search on CIFAR-100 and ImageNet is significantly reduced from a few GPU days to a few GPU hours of fine-tuning if a search policy is pretrained on CIFAR-10.

As the core motivation of NAS is automating the search, we offer a search procedure that can better generalize and transfer across different datasets instead of manually tuning the customized hyperparameters and optimizers to search on each individual dataset.

\begin{table}[t]
    \centering
    \caption{Highest validation accuracies measured during PC-DARTS supernet training. These values become $Acc(Env)$ for Equation~\ref{eq:transfer_reward} for transferability experiments.}
    \scalebox{0.9}{
    \begin{tabular}{l|c|c}
    \toprule
    \textbf{Dataset} & \textbf{Max Accuracy [\%]} & \textbf{GPU Days} \\ \midrule
    CIFAR-10 & 82.316 & ~3 \\
    CIFAR-100 & 56.380 & ~3\\
    ImageNet-32-12 & 48.765 & ~3.5 \\
    \bottomrule
    \end{tabular}
    }
    \vspace{-4mm}
    \label{tab:os_acc}
\end{table}

%% file: src/related.tex
\section{Related Work}
\label{sec:related}

DARTS~\cite{liu2018DARTS} proposes differentiable architecture search and has given rise to many follow-up works including P-DARTS~\cite{chen2019progressive} which breaks the search procedure into different stages and \cite{zela2019understanding} which propose early stopping. PC-DARTS~\cite{xu2020pcdarts} uses partial channel connections to decrease the memory cost in backpropagation and reduce bias toward parameterless operations.

However, recent works suggest that gradient-based NAS methods suffer from weaknesses such as the discretization error~\citep{yu2019evaluating, xie2020weight}, low reproducibility~\citep{yu2019evaluating} and bias towards shallow architectures \citep{shu2019understanding}. As a result, many attempts have focused on correcting these flaws. 
\cite{chen2020stabilizing} reduces the discretization error by forcing the supernet weights to generalize to a broader range of architecture parameters. \cite{dong2019GDAS} and \cite{xie2018SNAS} help bridge the optimization gap by using Gumbel-Softmax to sample a single operation per edge. ISTA-NAS~\citep{yang2020ista} recast NAS as a sparse coding problem in order to perform search and evaluation in the same setting. To improve generalizability, \cite{he2020milenas} reformulate DARTS as a 
mixed-level optimization problem. GAEA~\citep{li2020geometry} uses a simplex projection to ensure better convergence.

In comparison, we learn an agent in an actor-critic framework, which can be generalized and transferred across different datasets as a search policy. 
The RL-based training setup also allows for improved reproducibility and a natural incorporation of various exploration strategies.
Our supernet training uses random sampling inspired by \cite{li2019RS} to reduce the discretization error and bias on architecture depth.
Our experiments on transferability are partially inspired by \cite{wang2020gcn}, who use an RL agent to search for scalable analog circuit designs.

Finally, a number of RL-based NAS methods have been proposed. 
\cite{zoph2017NAS} use a controller trained by REINFORCE~\cite{Williams92Reinforce} to select architecture hyperparameters. ENAS~\citep{pham2018ENAS} is the first reinforcement learning scheme in weight-sharing NAS. \cite{bender2020can} show that guided policies exceed the performance of random search on vast search spaces. \cite{wang2018neural} uses Monte Carlo Tree Search to balance exploration with exploitation. These methods operate in a discrete space where the RL state keeps track of the partial architecture that is built over many steps. By contrast, L$^2$NAS selects an entire architecture per step and uses the RL state to keep track of the distribution of top performing architectures. 

\begin{table}[t]
	\centering
	\caption{Transferability results for L$^2$NAS for CIFAR-100 and ImageNet. `Direct' means directly searching on CIFAR-100 and ImageNet using the procedure in Section~\ref{sec:dartsEvaluation}.} 
	\label{table:transferability}
	\scalebox{0.9}{
	\begin{tabular}{l|cc|c}
		\toprule
		\textbf{CIFAR-100} & \textbf{Test Acc.  [\%]} & \textbf{Params [M]} & \textbf{GPU days} \\ \midrule
		DARTS 1st & 82.37 & 3.3 & 1.5 \\
		DARTS 2nd & 82.65 & 3.3 & 4.0 \\
		L$^2$NAS-Direct & 82.24 $\pm$ 0.19 & 3.5 & 1.0 \\
		L$^2$NAS-Transfer & 82.97 $\pm$ 0.29 & 4.0 & 0.1 \\ \midrule
		\textbf{ImageNet} & \textbf{Top-1/Top-5  [\%]} & \textbf{Params [M]} & \textbf{Search Cost} \\ \midrule
		DARTS 2nd & 73.1/91.3 & 4.7 & 4.0 \\
		L$^2$NAS-Direct & 74.8/92.2 & 4.9 & 1.0 \\
		L$^2$NAS-Transfer & 75.4/92.5 & 5.4 & 0.1 \\ 
		\bottomrule
	\end{tabular}
	}
	\vspace{-2mm}
\end{table}

%% file: src/conclusion.tex
\section{Conclusion}
\label{sec:conclusion}

In this paper, we propose L$^2$NAS to learn an optimizer for architecture hyperparameters in neural architecture search. Through an actor network taking feedback from the search history, L$^2$NAS produces continuous architecture hyperparameters $\alpha$, that are mapped into discrete architectures to obtain rewards. We propose a learning procedure for L$^2$NAS based on the continuous RL framework and a quantile-driven loss. L$^2$NAS learns to explore a large search space efficiently and achieves fast convergence to high-performing architectures. Experiments show that L$^2$NAS achieves state-of-the-art results on NAS-Bench-201 after querying only 1000 architectures. When working in the DARTS or OFA search space, L$^2$NAS produces architectures that achieve state-of-the-art test accuracies on CIFAR-10 and ImageNet  compared to a wide range of algorithms that exactly operate on the same search spaces. With transferability tests, we also demonstrate that an L$^2$NAS search policy pre-trained on CIFAR-10, can be used for effective search on CIFAR-100 and ImageNet with a low fine-tuning cost.